\documentclass[twoside,11pt]{article}

\usepackage{arxiv}

\usepackage{xspace}
\usepackage{courier}

\def\ie{\emph{i.e.,~}}
\def\eg{\emph{e.g.,~}}

\newcommand{\appname}{{\bf ktrain}\xspace}

\usepackage{balance}
\usepackage{graphicx}
\usepackage{caption}
\usepackage[list=true]{subcaption}

\usepackage{fancyvrb}

\usepackage{listings}
\usepackage{color}
\usepackage{textcomp}
\definecolor{listinggray}{gray}{0.9}
\definecolor{lbcolor}{rgb}{0.9,0.9,0.9}
\newcounter{Lcount}
\newcommand{\numsquishlist}{
  \begin{list}{\arabic{Lcount}. }
   { \usecounter{Lcount}
 \setlength{\itemsep}{0pt}      \setlength{\parsep}{3pt}
     \setlength{\topsep}{3pt}       \setlength{\partopsep}{0pt}
     \setlength{\leftmargin}{2.7em} \setlength{\labelwidth}{1em}
     \setlength{\labelsep}{0.5em} } }
\newcommand{\numsquishend}{\end{list}}

\newcommand{\squishlist}{
  \begin{list}{$\bullet$}
   { \setlength{\itemsep}{0pt}      \setlength{\parsep}{3pt}
     \setlength{\topsep}{3pt}       \setlength{\partopsep}{0pt}
     \setlength{\leftmargin}{1.5em} \setlength{\labelwidth}{1em}
     \setlength{\labelsep}{0.5em} } }
\newcommand{\squishend}{\end{list}}
\jmlrheading{A. S. Maiya}

\ShortHeadings{ktrain: A Low-Code Library for Augmented Machine Learning}{A. S. Maiya}
\firstpageno{1}

\begin{document}

\title{ktrain: A Low-Code Library for\\Augmented Machine Learning}

\author{\name Arun S. Maiya \email amaiya@ida.org \\
       \addr Institute for Defense Analyses \\Alexandria, VA, USA}

\maketitle

\begin{abstract}%   <- trailing '%' for backward compatibility of .sty file
We present \appname, a low-code Python library that makes machine learning more accessible and easier to apply. As a wrapper to TensorFlow and many other libraries (\eg \texttt{transformers}, \texttt{scikit-learn}, \texttt{stellargraph}), it is designed to make sophisticated, state-of-the-art machine learning models simple to build, train, inspect, and apply by both beginners and experienced practitioners.  Featuring modules that support \texttt{text} data (\eg text classification, sequence tagging, open-domain question-answering), \texttt{vision} data (\eg image classification),  \texttt{graph} data (\eg node classification, link prediction), and \texttt{tabular} data, \appname presents a simple unified interface enabling one to quickly solve a wide range of tasks in as little as three or four ``commands'' or lines of code.

\end{abstract}

\section{Introduction}  \label{introduction}
Machine learning workflows can be quite involved and challenging for newcomers to master. Consider the following steps.
~\\
~\\
\noindent
{\bf 1) Model-Building.}  The training data may reside in a number of different formats from files in folders to CSVs or {\em pandas} dataframes. If the data is large, it must be wrapped in a generator. Data must be preprocessed in specific ways depending on different factors such as the language of training texts (\eg English vs. Chinese) and whether or not transfer learning is being employed.  Learning rates, learning rate schedules, number of epochs, weight decay, and many other hyperparameters and settings must be selected or implemented.

\noindent
{\bf 2) Model-Inspection.} Once trained, a model is inspected in terms of both its successes and failures. This may include classification reports on validation performance, easily identifying examples that the model is getting the most wrong, and Explainable AI methods to understand why mistakes were made.

\noindent
{\bf 3) Model-Application.} Both the model and the potentially complex set of steps required to preprocess raw data into the format expected by the model must be easily saved, transferred to, and executed on new data in a production environment.  

\appname is a Python library for machine learning with the goal of presenting a simple, unified interface to easily perform the above steps regardless of the type of data (\eg text vs. images vs. graphs). Moreover,  each of the three steps above can be accomplished in as little as three or four lines of code, which we refer to as ``low-code'' machine learning.   \appname can be used with any machine learning model implemented in TensorFlow Keras (\Verb^tf.keras^). In addition, \appname currently includes out-of-the-box support for the following data types and tasks:
\clearpage
{\bf {\sc Text} Data:}
%\vspace{-2mm}
%\begin{itemize}
\squishlist
\item {\bf Text Classification}: auto-categorize documents across different dimensions
\item {\bf Text Regression}: predict numerical values (\eg prices) from textual descriptions
\item {\bf Sequence Tagging}: extract sequences of words that represent some concept of interest (\eg Named Entity Recognition or NER)
\item {\bf Unsupervised Topic Modeling}: discover latent themes buried in large document sets
\item {\bf Document Similarity with One-Class Learning}: find and score new documents based on thematic similarity to a set of seed documents
\item {\bf Document Recommendation}: recommend or return documents that are semantically-related to given text (\ie semantic search)
\item {\bf Text Summarization}: generate short summaries of long documents
\item {\bf Open-Domain Question-Answering}: submit questions to a large text corpus and receive exact answers
%\end{itemize}
\squishend
{\bf {\sc Vision} Data:}
\squishlist
\item {\bf Image Classification}: auto-categorize images across various dimensions
\item {\bf Image Regression}: predict numerical values (\eg age of person) from photos
\squishend
{\bf {\sc Graph} Data:}
%\begin{itemize}
\squishlist
\item {\bf Node Classification}: auto-categorize nodes in a graph (\eg social media accounts)
\item {\bf Link Prediction}: predict missing links in social networks (\eg friend suggestions)
%\end{itemize}
\squishend
{\bf {\sc Tabular} Data:} classification and regression (and causal inference) on data stored in tables

Many of the tasks above allow users to either choose from a menu of state-of-the-art models or employ a custom model.  With respect to text classification, for example, available models include cutting-edge Transformer models like BERT \citep{devlin2018bert, wolf2019huggingfaces} in addition to fast models such as fastText \citep{arm2016bag} and NBSVM \citep{wang2012baselines} that are amenable to being trained on a standard laptop CPU.
Other features include a learning-rate-finder to estimate an optimal learning rate \citep{smith2018disciplined}, easy-to-access learning rate schedules like the {\bf 1cycle policy} \citep{smith2018disciplined} and Stochastic Gradient Descent with Restarts (SGDR) \citep{loshchilov2016sgdr}, state-of-the-art optimizers like AdamW \citep{loshchilov2017decoupled}, ability to easily inspect classifications through Explainable AI and other methods, and a simple prediction API for use in deployment scenarios.
\appname is also bundled with pretrained, ready-to-use NER models for English, Chinese, and Russian. \appname is open-source, free to use under a permissive Apache license, and available on GitHub at: \url{https://github.com/amaiya/ktrain}.  In the next section, we compare and contrast our work with AutoML approaches.

\section{Augmented ML}  \label{augmented_ml}

Automatic machine learning (AutoML) solutions typically place a strong emphasis on automating subsets of the model-building process such as architecture search and model selection \citep{he2019automl}.  By contrast, \appname places less emphasis on this aspect of automation and instead focuses on either partially or fully automating other aspects of the machine learning (ML) workflow. For these reasons, \appname is less of a traditional AutoML platform and more of what might be called a ``low-code'' ML platform.  Through automation or semi-automation, \appname facilitates the full machine learning workflow from curating and preprocessing inputs (\ie ground-truth-labeled training data) to training, tuning, troubleshooting, and applying models. In this way, \appname is well-suited for domain experts who may have less experience with machine learning and software coding.  Where possible, \appname automates (either
algorithmically or through setting well-performing defaults), but also allows users to make choices that best fit their unique application requirements. In this way, \appname uses automation to augment and complement human engineers rather than attempting to entirely replace them. In doing so, the strengths of both are better exploited. Following inspiration from a blog post\footnote{\url{https://www.fast.ai/2018/07/16/auto-ml2/}} by Rachel Thomas of \verb^fast.ai^ \citep{howard2020fastai}, we refer to this as {\em Augmented Machine Learning} or AugML. For the remainder of this short paper, we will provide code examples to demonstrate ease-of-use.

\section{Building Models}  \label{building_models}

Supervised learning tasks in \appname follow a standard, easy-to-use template, which we now describe.

\noindent
{\bf STEP 1: Load and Preprocess Data.}  This step involves loading data from different sources and preprocessing it in a way that is expected by the model.  In the case of text, this may involve language-specific preprocessing (\eg tokenization).  In the case of images, this may involve auto-normalizing pixel values in a way that a chosen model expects.  In the case of graphs, this may involve compiling attributes of nodes and links in the network \citep{StellarGraph}. All preprocessing methods in \appname return a \verb^Preprocessor^ instance that encapsulates all the preprocessing steps for a particular task, which can be employed when using the model to make predictions on new, unseen data.
~\\
~\\
\noindent
{\bf STEP 2: Create Model.}  Users can create and customize their own model using \texttt{tf.keras} or select a pre-canned model with well-chosen defaults (\eg pretrained BERT text classifier \citep{devlin2018bert}, models for sequence tagging \citep{lample2016neural}, pretrained Residual Networks \citep{he2015deep} for image classification).  In the latter case, the model is automatically configured by inspecting the data (\eg number of classes, multilabel vs. multi-classification).  At this stage, both the model and the datasets are wrapped in a \texttt{ktrain.Learner} instance, which is an abstraction to facilitate training. 
~\\
~\\
\noindent
{\bf STEP 3: Estimate Learning Rate.}  Users can employ the use of a learning rate range test \citep{smith2018disciplined} to estimate the optimal learning rate given the model and data. Some models like BERT have default learning rates that work well, so this step is optional.
~\\
~\\
\noindent
{\bf STEP 4: Train Model.}  The \appname package allows one to easily try different learning rate schedules.  For instance, the \verb^fit_onecycle^  method employs a 1cycle policy \citep{smith2018disciplined}.  The \verb^autofit^  method employs a triangular learning rate schedule \citep{smith2018disciplined} with automatic early stopping and reduction of maximal learning rate upon plateau. Thus, specifying the number of epochs is optional in \verb^autofit^. The \verb^fit^ method, when supplied with the \verb^cycle_len^ parameter, decays the learning rate each cycle using cosine annealing.  Users can easily experiment with what works best for a particular problem. 

To illustrate ease of use, we provide fully-complete examples for two different tasks. 

\subsection{Example: Text Classification} \label{building_models.text_classification}
The first example is Chinese text classification.  More specifically, we train a Chinese-language sentiment-analyzer on a dataset of hotel reviews.\footnote{\url{https://github.com/Tony607/Chinese_sentiment_analysis}} 

~\\
\begin{scriptsize}
{\bf Fine-Tuning a BERT Text Classifier for Chinese:}
\begin{lstlisting}[language=Python]
import ktrain
from ktrain text as txt
# STEP 1: load and preprocess data
trn, val, preproc = txt.texts_from_folder('ChnSentiCorp', maxlen=75, 
                                         preprocess_mode='bert')
# STEP 2: load model and wrap in Learner
model = txt.text_classifier('bert', trn, preproc=preproc)
learner = ktrain.get_learner(model,train_data=trn, val_data=val)
# STEP 3: estimate learning rate 
learner.lr_find(show_plot=True) 
# STEP 4: train model
learner.fit_onecycle(2e-5, 4) 
\end{lstlisting}
\end{scriptsize}
Notice here that there is nothing special we need to do to support Chinese versus other languages like English.  The language and character encoding are auto-detected and processing proceeds accordingly.  Moreover, models are configured automatically through data inspection.   For instance, the data is automatically analyzed to determine the number of categories, whether or not categories are mutually-exclusive or not, and if targets are numerical or categorical.  The model is then auto-configured appropriately. 

\subsection{Example: Image Classification} \label{building_models.image_classification}

In the next example, we build an image classifier on the {\em Dogs vs. Cats} dataset\footnote{\url{https://www.kaggle.com/c/dogs-vs-cats}} with a standard ResNet50 model pretrained on ImageNet. As you can see in the code example below, the steps are very similar to the previous text classification example despite the task being completely different. 
~\\
~\\
\noindent
\begin{scriptsize}
{\bf Fine-Tuning a Pretrained ResNet50 Image Classifier:}
\begin{lstlisting}[language=Python]
import ktrain
from ktrain import vision as vis
# STEP 1: load and preprocess data
data_aug = vis.get_data_aug(horizontal_flip=True)
(trn,val,preproc)=vis.images_from_folder(datadir='data/dogscats',data_aug=data_aug,
                                             train_test_names=['train', 'valid'])                        
# STEP 2: load model and wrap in Learner
model = vis.image_classifier('pretrained_resnet50', trn, val, freeze_layers=15)
learner = ktrain.get_learner(model=model,train_data=trn,val_data=val,batch_size=64)                 
# STEP 3: find good learning rate
learner.lr_find(show_plot=True)         
# STEP 4: train
learner.autofit(1e-4) 
\end{lstlisting}
\end{scriptsize}

A unified interface to different and disparate machine learning tasks reduces cognitive load and allows users to focus on more important tasks that may require domain expertise or are less amenable to automation.

\section{Evaluating and Applying Models}  \label{evaluating_and_applying_models}
Once a model is trained, we would like to evaluate how well it learned what it was supposed to learn.  \appname provides a simple interface to perform various analyses to this end. To compute detailed validation (or test) metrics, the \verb^evaluate^ method can be invoked.  We can also easily identify the examples that the model got the most wrong by viewing examples with the highest validation (or test) loss using \verb^view_top_losses^:
~\\
~\\
\noindent
\begin{scriptsize}
{\bf Evaluating and Inspecting Models:}
\begin{lstlisting}[language=Python]
# compute validation metrics, confusion matrices, and show report
learner.evaluate() 
# example output of evaluate (uses validation set by default)
#                        precision    recall  f1-score   
#
#           alt.atheism       0.92      0.93      0.93       
#         comp.graphics       0.97      0.97      0.97      
#               sci.med       0.97      0.95      0.96       
#soc.religion.christian       0.96      0.96      0.96       
#
#              accuracy                           0.96      
#             macro avg       0.95      0.96      0.95      
#          weighted avg       0.96      0.96      0.96 

# view validation examples with highest loss
learner.view_top_losses()
\end{lstlisting}
\end{scriptsize}
\appname also features a simple and easy-to-use prediction API to make predictions on new and unseen examples.  A \verb^Predictor^ instance encapsulates both the model (\ie the underlying \verb^tf.keras^ model) and the preprocessing steps (\ie a \verb^Preprocessor^ instance) required to transform raw data into the format expected by the model.  The \verb^Predictor^ instance can easily be saved and reloaded for deployments to production environments.
\begin{scriptsize}

~\\
\noindent
{\bf Making Predictions on New Data:}
\begin{lstlisting}[language=Python]
predictor = ktrain.get_predictor(learner.model, preproc) # create predictor
predictor.predict(raw_data)                              # make predictions
predictor.save('/tmp/mypredictor')                       # save predictor
predictor = ktrain.load_predictor('/tmp/mypredictor')    # reload predictor
\end{lstlisting}
\end{scriptsize}

For a subset of tasks like text classification and image classification, \verb^Predictor^ instances expose an \verb^explain^ method that will attempt to {\em explain} how a model arrived at a decision for a particular example: \verb^predictor.explain(raw_data)^.  This can shed light on why certain decisions were successfully or unsuccessfully made by the model.  Explainable AI in \appname is powered by libraries such as {\bf shap} \citep{NIPS2017_7062} and {\bf eli5} with {\bf lime} \citep{ribeiro2016i}.

\section{Non-Supervised ML Tasks}  \label{other_examples}

All the examples covered thus far involve {\em supervised} machine learning.  Other tasks such as training {\em unsupervised} topic models to discover latent themes in document sets or using {\em pretrained} NER models follow slightly different steps than those described previously.  Despite involving a different pipeline, these non-supervised tasks also employ a low-code API and can be implemented in as little as three lines of code.  To illustrate this, we provide a code example for a fully-functional, end-to-end, {\bf open-domain question-answering system} using the well-studied {\em 20 Newsgroups} dataset.\footnote{\url{http://archive.ics.uci.edu/ml/datasets/Twenty+Newsgroups}}  We will first load the dataset into a Python list called \verb^docs^ using \verb^scikit-learn^ \citep{scikit-learn} (see Appendix \ref{appendix_a}). The basic idea here is to use the document set as a knowledge base that can be issued natural language questions to receive exact answers. In this case, we would like to issue questions about the subject matter buried in the {\em 20 Newsgroups} dataset and receive exact answers.  To accomplish this, the following steps are performed:

\numsquishlist
\item Index documents to a search engine.
\item Use the search index to locate documents that contain words in the question.
\item Extract paragraphs from these documents for use as contexts and use a BERT model pretrained on the SQuAD dataset to parse out candidate answers.
\item Sort and prune candidate answers by confidence scores and returns results.
\numsquishend

This entire workflow to build an end-to-end, open-domain question-answering (QA) system can be implemented with a surprisingly minimal amount of code with \appname:
~\\
~\\
\begin{scriptsize}
{\bf Building an End-to-End Open-Domain QA System in \appname}
\begin{lstlisting}[language=Python]
# build open-domain QA system
from ktrain.text.qa import SimpleQA
SimpleQA.initialize_index('/tmp/myindex') 
SimpleQA.index_from_list(docs, '/tmp/myindex', commit_every=len(docs))
qa = SimpleQA('/tmp/myindex')

# ask a question 
qa.ask('When did the Cassini probe launch?') # returns "October of 1997"
\end{lstlisting}
\end{scriptsize}

As shown above, upon building the QA system in {\bf only 3 lines of code}, we can submit natural language questions and receive exact answers.  In the example shown, we use the \verb^ask^ method to  submit the question, ``{\bf When did the Cassini probe launch?}''.  The candidate answer with the highest confidence score returned by the \verb^ask^ method is the correct answer of ``{\bf October of 1997}'' (see Appendix \ref{appendix_b}). Note that, for document sets that are too large to fit into a Python list, one can index documents using \verb^index_from_folder^ instead of \verb^index_from_list^. See Appendix \ref{appendix_c} for some additional low-code ML examples.

\section{Conclusion}

This work presented \appname, a low-code platform for machine learning. \appname currently includes out-of-the-box support for training models on \verb^text^, \verb^vision^, \verb^graph^, and \verb^tabular^ data.  As a simple wrapper to TensorFlow Keras, it is also sufficiently flexible for use with custom models and data formats, as well.  Inspired by other low-code (and no-code) open-source ML libraries such as \verb^fastai^ \citep{howard2020fastai} and \verb^ludwig^ \citep{molino2019ludwig}, \appname is intended to help further democratize machine learning by enabling beginners and domain experts with minimal programming or data science experience to build sophisticated machine learning models with minimal coding.  It is also a useful toolbox for experienced practitioners needing to rapidly prototype deep learning solutions.

\begin{footnotesize}
\vskip 0.2in
\bibliography{main}
\end{footnotesize}

% Appendix goes to a new page

\newpage

\appendix
\section{Loading the 20 Newsgroups Dataset} \label{appendix_a}

For the open-domain question-answering code example, we load the {\em 20 Newsgroups Dataset} using \verb^scikit-learn^.
~\\
\begin{lstlisting}[language=Python]
# load 20newsgroups dataset into a Python list
from sklearn.datasets import fetch_20newsgroups
remove = ('headers', 'footers', 'quotes')
newsgroups_train = fetch_20newsgroups(subset='train', remove=remove)
newsgroups_test = fetch_20newsgroups(subset='test', remove=remove)
docs = newsgroups_train.data +  newsgroups_test.data
\end{lstlisting}

\section{Open-Domain Question-Answering Example} \label{appendix_b}

We include a screenshot of a Jupyter notebook showing results from the question-answering API in \verb^ktrain^.

~\\
\includegraphics[scale=0.45]{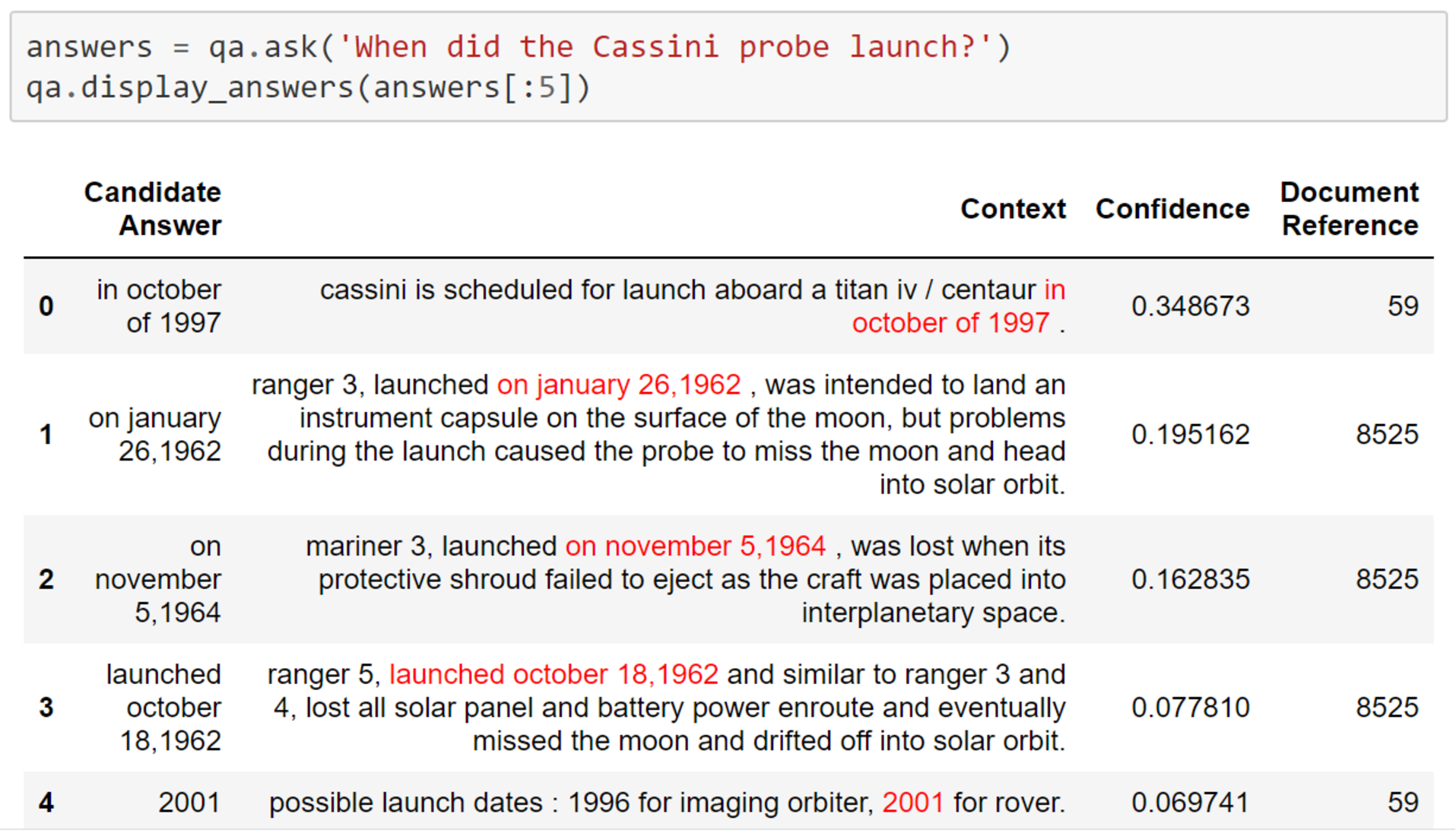}

~\\
We use the \Verb^qa.display^ method to format and display answers within a Jupyter notebook. The top candidate answer indicates that the Cassini space probe was launched in {\bf October of 1997}, which is correct. The specific answer within its context is highlighted in red under the column {\bf Context}. Since we used \Verb^index_from_list^ to index documents, the last column (populated from the \Verb^reference^ field in the returned \Verb^answers^ dictionaries) shows the list index associated with the newsgroup posting containing the answer. This \Verb^reference^ field can be used to peruse the entire document containing the answer with \Verb^print(docs[59])^. If using \Verb^index_from_folder^ to index documents, then the \Verb^reference^ field will be populated with the relative file path of the document instead.

\section{Additional Low-Code ML Examples} \label{appendix_c}

In this final section, we include some additional low-code examples of both supervised and non-supervised machine learning tasks in \appname to further illustrate ease-of-use.

~\\
\noindent
{\bf Named Entity Recognition with BioBERT embeddings:}
\begin{lstlisting}[language=Python]
import ktrain
from ktrain import text as txt
x_train= [['IL-2', 'responsiveness', 'requires', ...], ...]
y_train=[['B-protein', 'O', 'O', ...], ...]
(trn, val, preproc) = txt.entities_from_array(x_train, y_train)
model = txt.sequence_tagger('bilstm-bert', preproc, 
                             bert_model='monologg/biobert_v1.1_pubmed')
learner = ktrain.get_learner(model, train_data=trn, val_data=val,batch_size=128)
learner.fit(0.01, 1, cycle_len=5) # decays lr with cosine annealing
\end{lstlisting}

~\\
\noindent
{\bf Node Classification with Graph Neural Networks:}
\begin{lstlisting}[language=Python]
import ktrain
from ktrain import graph as gr
# STEP 1: load and preprocess data
(trn, val, preproc) = gr.graph_nodes_from_csv('cora.content','cora.cites',sep='\t')
# STEP 2: load model and wrap in Learner
model = gr.graph_node_classifier('graphsage', trn)
learner = ktrain.get_learner(model, train_data=trn, val_data=val batch_size=64)
# STEP 3: estimate learning rate
learner.lr_find(max_epochs=50, show_plot=True)
# STEP 4: train model
learner.autofit(0.01) # triangular lr schedule with early stopping
\end{lstlisting}

~\\
\noindent
{\bf Theme Discovery:} (no labeled training examples required)
\begin{lstlisting}[language=Python]
import ktrain
tm = ktrain.text.eda.get_topic_model(docs, n_features=10000) 
tm.build(docs, threshold=0.25) # documents to semantically meaningful vectors
tm.print_topics(show_counts=True) # prints discovered topics
tm.train_recommender()            # trains Nearest Neighbors model
rawtext = "Elon Musk leads Space Exploration Technologies (SpaceX), where he..."
tm.recommend(text=rawtext, n=5) # top 5 suggestions based on thematic similarity
\end{lstlisting}

~\\
\noindent
{\bf Zero-Shot Topic Classification:} (no labeled training examples required)
\begin{lstlisting}[language=Python]
from ktrain.text.zsl import ZeroShotClassifier
zsl = ZeroShotClassifier()
labels=['politics', 'elections', 'sports', 'films', 'television']
doc = "I am unhappy with decisions of government and will vote in 2020." 
zsl.predict(doc, labels=labels, include_labels=True)
# output:
# [('politics', 0.9829), 'elections', 0.9881), 
# ('sports', 0.0003), ('films', 0.0008), ('television', 0.0004)]
\end{lstlisting}
\end{document}